\documentclass{article}

\PassOptionsToPackage{numbers,sort&compress}{natbib}
\usepackage[preprint]{neurips_2026}

\usepackage[utf8]{inputenc}
\usepackage{lmodern}
\usepackage{hyperref}
\usepackage{url}
\usepackage{booktabs}
\usepackage{amsfonts}
\usepackage{xcolor}

\usepackage{amsmath}
\usepackage{amssymb}
\usepackage{amsthm}
\usepackage{graphicx}
\usepackage{subcaption}
\usepackage{enumitem}
\usepackage{bm}

\newcommand{\Bm}{\bm{B}}
\newcommand{\Km}{\bm{K}}
\newcommand{\Iv}{\bm{I}}
\newcommand{\Lm}{\bm{L}}
\newcommand{\xv}{\bm{x}}
\newcommand{\yv}{\bm{y}}
\newcommand{\R}{\mathbb{R}}
\newcommand{\gp}{\mathcal{GP}}
\DeclareMathOperator{\Var}{Var}

\newtheorem{proposition}{Proposition}
\newtheorem{remark}{Remark}

\title{Pitfalls and Remedies for Multi-Task Bayesian Optimization}

\author{%
  Carl Hvarfner \\
  Meta \\
  \texttt{hvarfner@meta.com} \\
  \And
  Sam Daulton \\
  Meta \\
  \texttt{sdaulton@meta.com} \\
  \And
  Max Balandat \\
  Meta \\
  \texttt{balandat@meta.com} \\
  \And
  Eytan Bakshy \\
  Meta \\
  \texttt{ebakshy@meta.com} \\
}

\begin{document}
\maketitle

\begin{abstract}
Bayesian optimization routinely warm-starts a target experiment with data from
related source tasks, and the multi-task Gaussian process is the textbook
surrogate for the job. We revisit this default in a controlled setting and find
that it misestimates the cross-task correlation even in the simplest
non-trivial case, affinely related source and target tasks, where a working
transfer-learning method should obviously succeed. We trace the failure to two
independent structural mechanisms. Per-task standardization, the textbook fix
for the affine slice ambiguity, propagates a finite-sample alignment error into
the recovered correlation. The marginal likelihood itself identifies the
correlation only at a per-sample rate that a Gaussian process at non-overlapping
designs further dilutes. We propose three conservative remedies that follow
from the analysis: promoting per-task means and scales to model parameters,
restricting the task covariance to non-negative correlations, and co-locating
part of the source and target designs. Across synthetic multi-task problems and
surrogate-based hyperparameter-tuning transfer, these remedies recover the target-only
baseline on the simple instances, while the broader failure persists on harder
instances and across most rank-based and latent-context variants.
\end{abstract}

\section{Introduction}
\label{sec:intro}

Bayesian optimization (BO)~\citep{Frazier2018,Garnett2023,Shahriari2016,Snoek2012}
is a workhorse for sample-efficient experimentation, and in production settings,
target experiments rarely arrive in
isolation~\citep{Golovin2017,Perrone2018,Feurer2018}. BO transfer learning
(BOTL) is the default story whenever a target experiment has a related
predecessor: gather the source data, fit a multi-task Gaussian process (MTGP),
and expect fewer target evaluations to find the optimum. Reports on BOTL tend to
emphasize positive results on curated suites~\citep{Swersky2013,Bardenet2013,Feurer2018},
and library defaults~\citep{BoTorch,GPyTorch} present the Intrinsic
Coregionalization Model (ICM)-based MTGP as the go-to surrogate. Controlled
comparisons against target-only BO~\citep{Eggensperger2021,Pineda2021}, and
direct audits of whether the standard model recovers correct task correlations,
remain comparatively rare.

Yet on two affinely related tasks drawn from a standard benchmark function, the
textbook MTGP misestimates the cross-task correlation: it attenuates the
recovered correlation and, once more than one source is present, can even flip
its sign. Affinely related tasks are the textbook example a working
transfer-learning method must handle: the source is a perfect linear image of
the target, so every standard MTGP variant should recover near-perfect
correlations and transfer should obviously help. We find the opposite: similar
pathologies persist not only for the textbook ICM but for nearly every
multi-task and rank-based variant in common use, and across our multi-task BO
grid the textbook MTGP loses to a single-task Gaussian process (GP) on most base
functions -- the textbook signature of \emph{negative
transfer}~\citep{PanYang2010,Wang2019Negative}. A method failing here is failing
structurally, not because the task is hard; prior BOTL benchmarks report
ICM-based MTGPs on suites where the underlying transferability is itself
uncertain, so a clean failure on affinely related tasks is diagnostic. We trace
the failure to a structural identifiability defect in the standard
parameterization, not a fitting accident.

The failure has two structurally distinct, independent causes: a finite-sample
per-task \emph{standardization} error that rides the affine reparameterization
symmetry into the recovered correlation, and an information-theoretic floor on
\emph{correlation inference} that a GP at non-overlapping designs further
dilutes. Both bite at the budgets BOTL practice actually has
(Fig.~\ref{fig:correlation_seeds}); \S\ref{sec:pitfall} develops each in turn.

\paragraph{Our contributions} are the following:
\begin{itemize}[leftmargin=*]
\item \textbf{Two structural pitfalls of the textbook MTGP, with theory.} We isolate two
  independent mechanisms that make the textbook MTGP misestimate task correlations
  on affinely related tasks: a finite-sample per-task standardization error and
  an information-theoretic floor on correlation inference that a non-overlapping
  GP further dilutes.
\item \textbf{Three remedies that recover the target-only baseline on simple instances.} We identify per-task means
  and scales as model parameters, a non-negativity constraint on
  $\rho$, and co-located source/target queries as the configuration
  that minimizes the inference-side variance the two pitfalls leave on
  the table (\S\ref{subsec:pitfall_remedies}).
\item \textbf{Empirical demonstration.} We evaluate the textbook ICM,
  several MTGP variants, and the proposed remedies on simple
  semi-synthetic affine problems, on surrogate-based LCBench hyperparameter
  optimization (HPO), and on surrogate-based deep-learning (pd1) and LLM
  (\textsc{ifeval}~\citep{Chen2024Admire}) tuning benchmarks
  (\S\ref{sec:results}).
\end{itemize}

\section{Background}
\label{sec:background}

\paragraph{Bayesian optimization and transfer learning.}
BO addresses sequential black-box optimization of a noisy objective
$f: \mathcal{X} \to \R$ under a limited query budget: at each step a probabilistic
surrogate (typically a GP) and an acquisition function decide the next
$\xv \in \mathcal{X}$ to evaluate from past observations
$\mathcal{D}_t = \{(\xv_i, y_i)\}_{i=1}^{N_t}$. BOTL augments this with
$T-1$ source datasets $\{\mathcal{D}^{(s)}\}_{s=1}^{T-1}$ on related but
non-identical objectives $f_s$ from past or cheaper experiments; the
surrogate is fit jointly across source and target so that source
evaluations sharpen the target posterior whenever the underlying tasks
are correlated. The benefit hinges on whether the surrogate can both
(i) recover the cross-task correlation from finite source budgets and
(ii) align source and target onto a common scale; the failure modes
examined in this paper sit at exactly those two steps.

\paragraph{GP regression.} A GP~\citep{RasmussenWilliams2006,Stein1999} is a distribution
over $f : \mathcal{X} \to \R$ such that any finite set of evaluations is jointly Gaussian.
Given observations $\bm{y} = f(\bm{X}) + \bm{\varepsilon}$ with
$\bm{\varepsilon} \sim \mathcal{N}(0, \sigma^2_{\mathrm{noise}}\Iv)$ and a positive
definite kernel $k_x$, the posterior is again a GP, with predictive mean
$\bm{k}_*^\top(\Km + \sigma_{\mathrm{noise}}^2\Iv)^{-1}\yv$ and variance
$k_x(\xv_*,\xv_*) - \bm{k}_*^\top(\Km + \sigma_{\mathrm{noise}}^2\Iv)^{-1}\bm{k}_*$
at a test input $\xv_*$, where $\bm{k}_* = k_x(\xv_*, \bm{X})$ and $\Km$ is
the Gram matrix on the training inputs. Hyperparameters are typically fit by
maximizing the marginal log-likelihood (MLL).

\paragraph{MTGP and the ICM.} An MTGP~\citep{Bonilla2008} models $T$
correlated functions $\{f_t\}_{t=1}^T$ jointly. In the ICM
parameterization~\citep{GoulardVoltz1992}, the cross-task covariance is the Kronecker
product of a $T \times T$ task covariance $\Bm$ and a shared input kernel $k_x$:
$\operatorname{Cov}(f_t(\xv), f_{t'}(\xv')) = B_{tt'}\, k_x(\xv,\xv')$,
where the diagonal entry $B_{tt}$ is the per-task signal variance and the
off-diagonal $B_{tt'}$ is the cross-task covariance; the unitless
task-correlation matrix is $\rho_{tt'} = B_{tt'}/\sqrt{B_{tt}B_{t't'}}$.
Intuitively, $k_x$ controls smoothness within a task while $\Bm$ controls
how strongly information is shared across tasks; the model is \emph{separable}
in input and task. The MTGP additionally permits per-task observation noise
$\sigma_{\mathrm{noise},t}^2$ and per-task constant means $\mu_t$ -- the latter
possible in principle but seldom discussed or implemented in practice.

\paragraph{Task-correlation matrix and standardization.} The standard
$\Bm=\Lm\Lm^\top+\mathrm{diag}(\bm{v})$ parameterization~\citep{Bonilla2008}
is the default in mainstream libraries~\citep{BoTorch,GPyTorch}. Outputs are
typically standardized either globally (one $(\hat\mu,\hat\sigma)$ across all
tasks) or per task ($(\hat\mu_t,\hat\sigma_t)$ from each task's observations
alone). Standardization brings the outputs into the regime in which the GP's
default hyperparameter priors, centered on unit signal variance, are
informative -- the same normalized regime in which priors and initializations
for high-dimensional BO are calibrated~\citep{Hvarfner2024Vanilla,Papenmeier2022}
-- and is essentially required for stable MLL optimization on heterogeneously
scaled tasks.

\paragraph{The model's whitened space.} Standardization is effectively bi-level.
A first layer is the raw standardization of the outputs by an outcome transform
$(m_t, s_t)$, set to the empirical $(\hat\mu_t, \hat\sigma_t)$ under per-task
standardization. A second layer is the normalization the GP performs by learning
a per-task mean constant $c_t$ and an ICM diagonal $B_{tt}$. The signal the GP
treats as zero-mean, unit-variance per task is therefore the \emph{whitened}
signal $z_t(\xv) = ((y_t - m_t)/s_t - c_t)/\sqrt{B_{tt}}$. Learning the per-task
mean and signal is not sequential with fitting the remaining hyperparameters,
but it is natural to think of the correlation parameters and lengthscales as
learned in this whitened space: the off-diagonal $\rho_{tt'}$ encodes correlation
\emph{between whitened signals}, not raw outputs.

\section{Related Work}
\label{sec:related}

\paragraph{Pearson-based (correlation-inferring) MTGPs.} The Linear Model of Coregionalization (LMC)/rank-1 ICM
originates in geostatistics~\citep{Journel1978,GoulardVoltz1992,Wackernagel2003} and
entered the GP literature via~\citet{Bonilla2008}; multi-task BO
followed~\citep{Swersky2013,Poloczek2017,Bardenet2013}. The class spans free
and rank-restricted $\Bm$, latent-context embeddings, pooled single-task GPs
($\Bm=\bm 1\bm 1^\top$), and hierarchical priors on $\Bm$ -- all infer task
structure through pairwise Pearson covariance and inherit the affine
identifiability defect of \citet{Alvarez2012,AndersonRubin1956,LopesWest2004}
we sharpen below.

\paragraph{Rank-based ensembles.} Per-task GP ensembles combined by
rank-agreement weights, such as the ranking-weighted GP ensemble
(RGPE)~\citep{Feurer2018,Wistuba2018}, sidestep the joint task covariance
entirely and instead reweight independently fit per-task GPs.

\paragraph{Distribution-matching transforms.} A per-task copula
$\Phi^{-1}\!\circ\hat F_t$ absorbs any monotone per-task map, and a shared model
is fit in $z$-space~\citep{Salinas2020}; this is the basis of the Gaussian
Copula Process (GCP).

\paragraph{Other paradigms.} Offline meta-learning and amortized
policies~\citep{Wang2024,Volpp2020,Perrone2018} sidestep per-target fitting at
the cost of $1$--$2$ orders of magnitude more source data; pre-trained-GP work
moves the identifiability question to meta-train time.

\section{Two pitfalls of affine MTGPs}
\label{sec:pitfall}

We present our two structural pitfalls in the simplest imaginable transfer
setting: affinely related source and target tasks. Because the source is a
perfect linear image of the target -- the most generous case for transfer, with
true task correlation exactly $\pm 1$ -- any failure to recover the correlation
here is a structural defect rather than task difficulty.

\paragraph{Setup: affine source-target family.}
Let $f \sim \gp(0, k_x)$ be a latent function with normalized base kernel
$k_x(\xv, \xv) = 1$ and define $T$ tasks by
$y_t(\xv) = a_t\, f(\xv) + b_t + \varepsilon_t$,
where $a_t > 0$, $b_t \in \R$, and
$\varepsilon_t \sim \mathcal{N}(0, \sigma_{\mathrm{noise},t}^2)$
independently for $t = 1, \ldots, T$.
Assume the high signal-to-noise ratio (SNR) regime:
$\sigma_{\mathrm{noise},t}^2 \ll a_t^2$.
Consider an ICM model~\citep{GoulardVoltz1992} with task covariance
$\Bm \in \R^{T \times T}$ whose diagonal entries $B_{tt}$ are the per-task
signal variances ($B_{tt} = a_t^2$ at the truth), per-task constant means
$\mu_t$, and shared normalized base kernel $k_x$, so that the joint covariance is
\begin{equation}
\label{eq:icm_joint_cov}
    \operatorname{Cov}\bigl[y_t(\xv),\, y_{t'}(\xv')\bigr]
    = B_{tt'}\, k_x(\xv, \xv')
    + \sigma_{\mathrm{noise},t}^2\, \delta_{tt'}\, \delta_{\xv\xv'}.
\end{equation}
Identifying $\rho$, not $\Bm$, is the goal; pinning
$B_{\mathrm{target,target}}=1$ removes the residual scale ambiguity in the
target row/column.
While we develop the analysis on the affine source-target family for
tractability, the underlying mechanisms are model-agnostic and apply broadly
to any multi-task surrogate that estimates per-task scale and pairwise
covariance from finite samples.

As we will show, the textbook MTGP fails on affinely related tasks for two
structurally distinct, independent reasons. \emph{Aligning} the source carries a
finite-sample per-task standardization error that propagates into the recovered
correlation even when $\rho^\star=1$ (\S\ref{subsec:pitfall_standardize}).
\emph{Inferring} the correlation is information-bound: the data identifies the
task correlation poorly even when alignment is exact
(\S\ref{subsec:pitfall_correlation}). We close in
\S\ref{subsec:pitfall_remedies} with three configuration choices that shrink the
regime where either bites.

\begin{figure}[t]
\centering
\includegraphics[width=\linewidth]{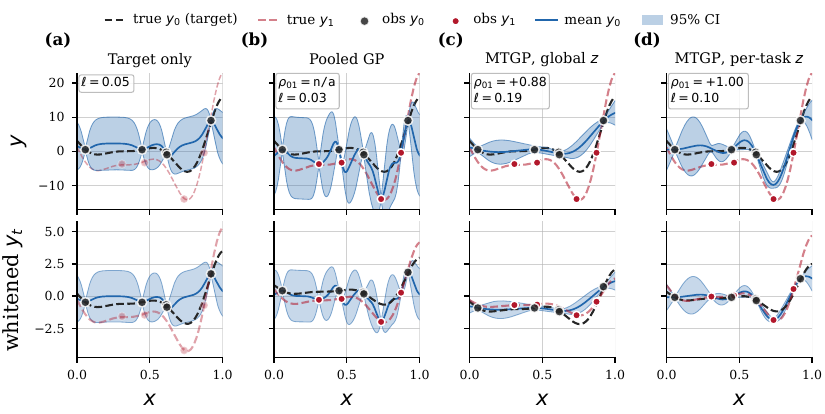}
\caption{\textbf{Four GP configurations on affinely related tasks.} Target task
$y_0 = f$ on the Forrester minimization benchmark and an affinely related source
$y_1 = 1.7\,f - 4$, with $N_0 = 5$ target points and $N_1 = 4$ source points.
\textbf{Top row}: GP posterior mean and $\pm 2\sigma$ bands over $y$; insets
report the learned correlation $\rho_{0,1}$ and the lengthscale $\ell$.
\textbf{Bottom row}: the same elements in the GP's whitened working space $z_t$,
where tasks that share affine structure should overlap. Panel~(c): the textbook
MTGP attenuates the recovered correlation to $+0.88$. Panel~(d): the recommended
configuration recovers the structure.}
\label{fig:affine_pitfall}
\end{figure}

Fig.~\ref{fig:affine_pitfall} makes the joint failure mode visible
on a 1D Forrester benchmark with two affinely related tasks: the textbook ICM
(panel~c) attenuates the recovered correlation $\hat\rho_{st}$ and pulls the
target posterior the wrong way, while the recommended configuration (panel~d) --
per-task means, free $\Bm$, per-task noise, and per-task standardization --
recovers the structure.

\subsection{Standardization is hard even when $\rho^\star=1$}
\label{subsec:pitfall_standardize}

Suppose the source and target are perfectly correlated. The MTGP still has to align the source
onto the target's scale by estimating per-task constants
$(\hat\mu_s,\hat\sigma_s)$ from the same $N_s$ source observations it
then transfers. The standardization estimates $(\hat\mu_s,\hat\sigma_s)$ themselves carry
finite-sample error of order $1/\sqrt{N_s}$. Because per-task standardization is
an affine reparameterization, this error transfers directly into the
source--target block of the task covariance $\Bm$, biasing the recovered
correlation $\hat\rho_{st}$ at the same $1/\sqrt{N_s}$ rate -- even when the
true correlation is exactly $1$. Per-task standardization plays a dual
role here: it is simultaneously the textbook fix for the affine slice
ambiguity and the source of the noise we are trying to remove.

\begin{proposition}[Standardization-error propagation under $\rho^\star=1$]
\label{prop:standardize_propagation}
Under the affine setup with $\rho^\star_{st}=1$, let
$(\hat\mu_s,\hat\sigma_s)$ be the empirical mean and standard
deviation of $N_s$ source observations and let
$(\hat\alpha_s,\hat\beta_s)=(1/\hat\sigma_s,\hat\mu_s)$ be the
per-task standardization map. Then
\[
   \hat\mu_s = \mu_s + \mathcal{N}\!\bigl(0,\tfrac{\sigma_s^2}{N_s}\bigr) + o_p(N_s^{-1/2}),
   \qquad
   \hat\sigma_s = \sigma_s + \mathcal{N}\!\bigl(0,\tfrac{\sigma_s^2}{2 N_s}\bigr) + o_p(N_s^{-1/2}),
\]
and the recovered source--target correlation under per-task
standardization satisfies
\[
   \hat\rho_{st}/\rho^\star_{st} \;=\; 1 \;+\; \mathcal{N}\!\bigl(0,\tfrac{1}{2 N_s}\bigr) \;+\; o_p(N_s^{-1/2}).
\]
Proof in App.~\ref{app:standardize_harm_proof}.
\end{proposition}

Notably, the relative error in $\hat\rho_{st}$ is governed by the smaller of the
two source-side sample sizes: doubling $N_s$ only halves the variance, so the
bias persists deep into the BO budget regime. The practical consequence is
visible in Fig.~\ref{fig:affine_pitfall}(c): incorrect offset estimation lowers
the recovered correlation and shrinks the effective influence of the source
task's best point on the target posterior. At the small source budgets BO
practice actually has, replacing the true per-task scale by its empirical
estimate collapses the recovered maximum-likelihood estimate (MLE) well below the
true correlation on otherwise identical data; the bias only fades once the source
budget grows by an order of magnitude or more. Standardization is the textbook
fix for the affine slice ambiguity, but at BO source budgets it transports the
same $1/\sqrt{N_s}$ noise into $\hat\rho_{st}$ that the slice was supposed to
remove.

\subsection{Inferring the task correlation is hard}
\label{subsec:pitfall_correlation}

Suppose now that alignment is given -- standardization is exact --
and the only unknown is the cross-task correlation $\rho$. The
per-sample Fisher information is bounded above by the
bivariate-Gaussian rate, which both Pearson and Spearman estimators
saturate up to a small constant; that rate then degrades further on
a GP fit at non-overlapping designs. The data identifies $\rho$
poorly even when everything else is given.

\begin{proposition}[Per-sample $\rho$-detection floor]
\label{prop:cr_bound}
For two tasks observed at $N$ shared design points under the affine setup, the
maximum-likelihood Pearson estimator satisfies
\[
  \Var[\hat\rho]\ge \frac{(1-\rho^2)^2}{N(1+\rho^2)},
\]
with equality in the noiseless limit. The Spearman rank estimator $\hat\rho_S$
inherits the same $\Theta(1/\sqrt N)$ per-sample standard error (SE) up to the
asymptotic relative efficiency (ARE) factor $9/\pi^2\approx 0.912$
\citep{Borkowf2002}. For mismatched designs $|X_s|=N_s\le|X_t|=N_t$ the smaller
task dominates the rate. Full proof and Spearman ARE in
App.~\ref{app:cr_bound_proof}.
\end{proposition}

\begin{proposition}[GP information dilution]
\label{prop:gp_dilution}
For shared input kernel $k_x$ on $\Omega\subset\R^d$ with correlation length
$\ell$, the Fisher information of an $N$-point GP fit obeys
\[
  I_N(\rho)\le N_{\mathrm{eff}}\,\frac{1+\rho^2}{(1-\rho^2)^2},
  \qquad
  N_{\mathrm{eff}}\le\min\{N,(\mathrm{diam}(\Omega)/\ell)^d\}:
\]
a GP identifies $\rho$ from at most $N_{\mathrm{eff}}$ effective paired
observations, strictly fewer than the $N$ a naive paired test would assume. Proof
in App.~\ref{app:gp_dilution_proof}.
\end{proposition}

\begin{remark}
\label{rem:dilution}
The dilution factor depends on $d$ and $\ell$ but not on $\rho$;
high-dimensional source designs at moderate lengthscale can be
arbitrarily worse than the iid paired-sample bound.
\end{remark}

\begin{figure}[t]
\centering
\includegraphics[width=\textwidth]{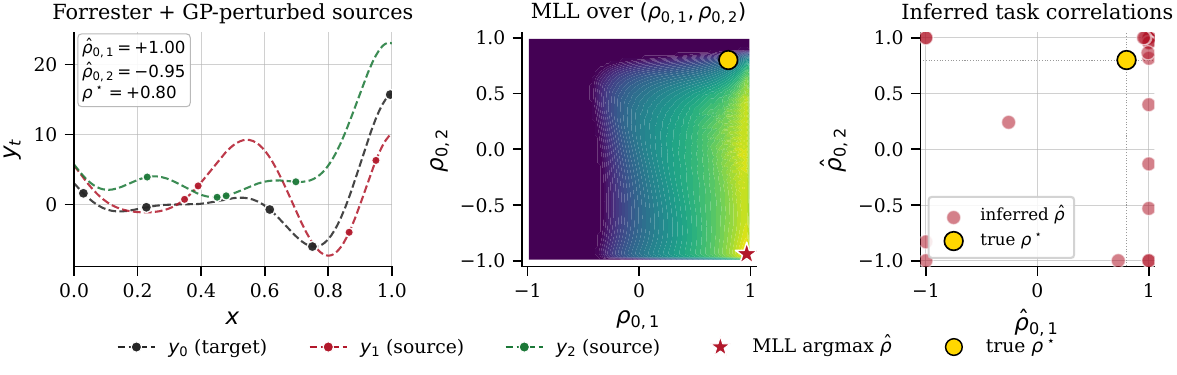}
\caption{\textbf{MTGP correlation inference at typical BO source budgets.}
1D Forrester target under minimization with two GP-perturbed source tasks
$g_t = f + h_t$, calibrated to a true target--source correlation
$\rho^\star = 0.8$ and fitted on $5{+}4{+}4$ stratified-uniform observations.
\textbf{Left:} a representative seed, where the MLE recovers
$\hat\rho_{0,1} = +1.00$ and $\hat\rho_{0,2} = -0.95$.
\textbf{Middle:} the MLL as a function of the two target--source correlations is
broad and ridge-like, so unlucky designs land the optimum almost anywhere along
that ridge.
\textbf{Right:} MLE estimates across $25$ independent seeds; sign flips and
corner saturation are common, and correlation inference at these budgets is
unreliable in our setting.}
\label{fig:correlation_seeds}
\end{figure}

Resolving a moderate correlation to useful precision requires far more paired
observations than BO source budgets provide: the Cram\'er--Rao (CR) standard
error is large at BO budgets, and the GP dilution above shrinks the effective
sample count further. As a result the MTGP struggles to tell a moderately useful
source from a useless one at the budgets BOTL practice actually has
(Fig.~\ref{fig:correlation_seeds}) -- the inference-side mechanism behind the
negative-transfer outcomes observed empirically in \S\ref{sec:results}.

\subsection{Three remedies}
\label{subsec:pitfall_remedies}

The analysis above motivates three conservative configuration choices. None is
novel; each is a practical necessity at the source budgets BO practice has.
Remedy~1 promotes the per-task offset and scale to model parameters, removing
the empirical-standardization noise isolated in
\S\ref{subsec:pitfall_standardize}. Remedies~2 and~3 then reduce the
correlation-inference variance that the bound of
Prop.~\ref{prop:cr_bound} and the GP dilution of Prop.~\ref{prop:gp_dilution}
together leave on the table.

\paragraph{Remedy 1: per-task means and per-task scales as model parameters.}
Two affinely related tasks differ in raw range but become identical up to noise
after per-task standardization; this visual identity is what tempts practitioners
to trust empirical standardization in the first place. Even small per-task
offsets -- the kind visible in Fig.~\ref{fig:affine_pitfall} -- nevertheless
destroy the model's correlation estimate when only empirical standardization is
used. Promoting the per-task offset and scale to model parameters $\mu_t, B_{tt}$,
fit jointly with the rest of the model, lets the marginal likelihood balance their
estimation against the data instead of paying the finite-sample alignment cost up
front. Per-task scales necessarily apply to the source tasks, though the target
task may still carry a fixed output scale. This remedy is not novel -- it is
well-known in the GP literature -- but it is important enough to state explicitly,
since the remaining remedies build on it.

\paragraph{Remedy 2: positive-correlation restriction.}
Joint inference of the task-covariance parameters is overparameterized at small
$N$, and the covariance entries are precisely the high-variance ones
(\S\ref{subsec:pitfall_correlation}); the positivity constraint is therefore a
practical necessity, not an innovation. Concretely, constrain $\rho_{st}\ge 0$
(e.g., parameterise $\Bm=\Lm\Lm^\top+\mathrm{diag}(\bm{v})$ with $\Lm$ entrywise
non-negative). Allowing free-form, signed correlations rests on faulty logic: a
meaningfully negatively related source is not something a practitioner would
realistically choose to include, yet leaving the sign free exposes the model to
the wrong-sign mode that finite-sample noise can flip the MLE into
(Fig.~\ref{fig:correlation_seeds}), and the CR floor of Prop.~\ref{prop:cr_bound}
leaves negatively correlated sources unidentifiable in any case. We show in the
results that this wrong-sign failure occurs in practice, and the appendix
correlation densities (App.\ Fig.\ \ref{fig:inferred_correlation}) make it
explicit.

\paragraph{Remedy 3: co-locating source and target observations.}
Each shared input does double duty: it lifts the Fisher information for $\rho$ to
the paired-test rate of Prop.~\ref{prop:cr_bound} that a non-overlapping GP fit
forfeits, and, when tasks are related, induces correlated bias across per-task
offset/scale estimators -- a variance-reduction effect that helps Remedy~2 too.
For correlation inference, re-evaluating \emph{any} source point on the target is
beneficial; re-evaluating a strong source point can additionally be desirable in
its own right. Co-locating does not by itself reduce input-space coverage of the
target, but it reduces \emph{overall} coverage across tasks when the tasks are in
fact highly correlated -- something we cannot count on in advance. This exposes a
trade-off between accurate correlation inference and the coverage that accurate
inference would earn. On real data, this remedy appears as the
source-design-overlap ICM variants on LCBench, whose effect on the inferred
correlation is shown in App.\ Fig.\ \ref{fig:inferred_correlation}
(\S\ref{subsec:lcbench}).

\section{Experiments}
\label{sec:results}

We stress-test the analysis on a multi-task BO grid covering both
synthetic test functions (\S\ref{subsec:synthetic}) and surrogate-based HPO transfer
on LCBench (\S\ref{subsec:lcbench}), pd1, and \textsc{ifeval}
(\S\ref{subsec:hpo}). Affinely related tasks are the
canonical case where transfer ``obviously should work'': the achievable upper
bound on performance is known by construction, since an oracle given the
per-task standardization parameters recovers the target with no transfer cost.
A method failing here is failing structurally, not because the task is hard;
prior BOTL benchmarks~\citep{Eggensperger2021,Pineda2021,Golovin2017} report
ICM-based MTGPs on suites where the underlying transferability is itself
uncertain, and few of them cleanly evaluate the model class on a setting where
transfer should obviously succeed. We benchmark against the implicit affine
ceiling rather than running the oracle directly.

We benchmark a Vanilla GP (target-only), three ICM variants -- ICM~(Shared),
ICM~(Positive), and ICM~(Free) -- and RGPE~\citep{Feurer2018}, all using
$q\text{LogNEI}$~\citep{Ament2024LogEI} and dimension-scaling lengthscale
priors~\citep{Hvarfner2024Vanilla}, alongside QuantileBO~\citep{Salinas2020}
and Adaptive Bayesian Linear Regression (ABLR)~\citep{Perrone2018}. The three ICM variants isolate the model-side
choices of \S\ref{subsec:pitfall_remedies}. \emph{ICM~(Shared)} is the textbook
configuration: global standardization, a single mean across tasks, and a free
task covariance $\Bm$. \emph{ICM~(Positive)} adds per-task means and per-task
scales (Remedy~1) and a non-negativity constraint on $\rho$ (Remedy~2).
\emph{ICM~(Free)} keeps per-task means and scales but drops the sign constraint,
allowing $\rho$ to be negative. QuantileBO is dropped on \textsc{ifeval}, where
its dimension-$19$ task-correlation matrix is not positive definite; two
source-design-overlap ICM variants (Remedy~3) appear only in the appendix
ablations on the synthetic grid and LCBench, since the initialization across
methods is otherwise not identical. The method-to-configuration map is
given in App.~\ref{app:tasks}.

\subsection{Synthetic test functions}
\label{subsec:synthetic}

Per replication we draw a positive-affine source family
$g_t(\xv)=a_t f(\xv)+b_t$ from the affine setup with $N_s=12$ observations per
source task. The target is initialized with Sobol points
($n_{\mathrm{init}} = d+1$); the full configuration is in
App.~\ref{app:task_synthetic} (per-method ablation in
Fig.~\ref{fig:synthetic_grid_abl}). Even the textbook ICM~(Shared) trails the
target-only Vanilla GP on the simple instances.

\begin{figure}[t]
\centering
\includegraphics[width=\textwidth]{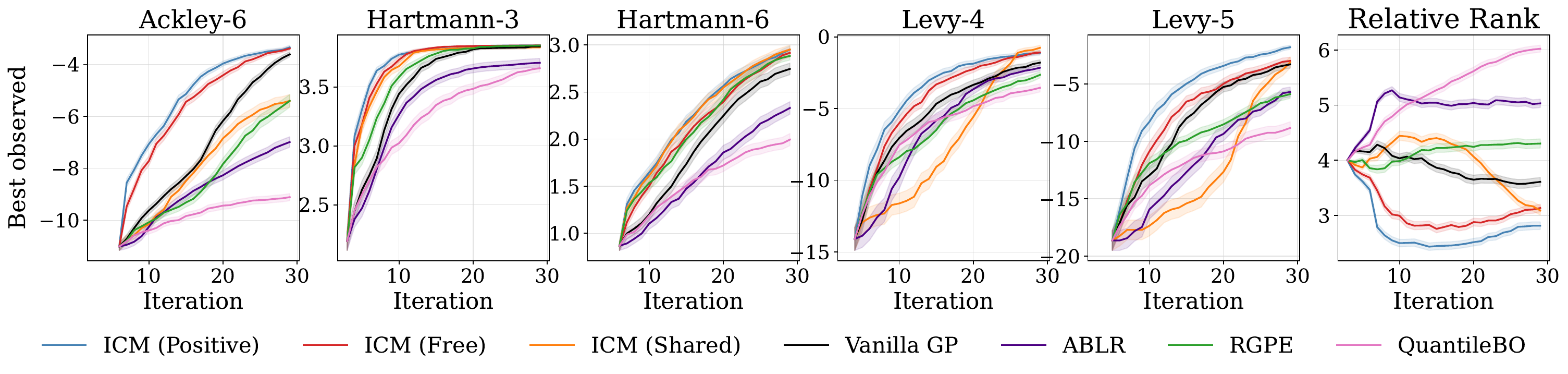}
\caption{\textbf{Synthetic affine-TL grid.} Best-observed target value
vs.\ acquisition iteration on Hartmann-3, Hartmann-6, Ackley-6, Levy-4, and Levy-5,
mean $\pm 1$~SE across $99$ seeds, with a final Relative Rank panel aggregating
rank across the five base functions (lower is better). The textbook ICM~(Shared)
trails the target-only Vanilla GP on the high-SNR Hartmann instances, the regime
where transfer should be easiest; the body (\S\ref{subsec:synthetic}) reads off
the full method ordering. HyperBO~\citep{Wang2024} is excluded as its design
regime ($\sim$24 tasks, hundreds of evaluations per task) lies far outside ours.}
\label{fig:synthetic_grid}
\end{figure}

Fig.~\ref{fig:synthetic_grid} matches the structural picture of
\S\ref{sec:pitfall} on the favorable simple instances, and we read it as a
minimum bar: a method that fails here cannot be trusted to transfer in the wild.
The simple synthetic experiments make the issues of the transfer methods
immediately visible. The textbook ICM~(Shared) struggles to beat the Vanilla GP
until the source budget grows large enough: global standardization is not in
itself harmful, but it leaves the model dependent on downstream hyperparameter
inference that is fragile at the budgets we test, and RGPE inherits the same
per-task standardization sensitivity through its base learners. More broadly, all
of these methods rely on aligning the tasks to some degree, and a
higher-complexity method (e.g.\ QuantileBO, ABLR) does not necessarily transfer
better than a simpler one. The per-task ICM variants (ICM~(Positive) and
ICM~(Free), which add Remedy~1 on top of the textbook configuration) recover the
Vanilla GP on the positive-affine instances; ICM~(Positive) additionally enforces
Remedy~2 and is the more robust of the two on a per-seed basis. The wrong-sign
mode is removed, but the Prop.~\ref{prop:cr_bound} floor remains, and on the
harder Ackley and Levy instances the gap to Vanilla narrows but does not close.

\subsection{Surrogate-based transfer: LCBench HPO}
\label{subsec:lcbench}

LCBench~\citep{Zimmer2021} provides HPO tasks over a shared $7$-D neural-network
configuration space. We use it as a semi-synthetic testbed: each target is a
Gaussian-process surrogate fitted to a real LCBench dataset, and its sources are
small affine transformations of that same surrogate, so the underlying task
structure is positive-affine by construction. Each target is warm-started by
\textbf{two positively correlated LCBench sources} (full setup in
App.~\ref{app:task_lcbench}).

\begin{figure}[t]
\centering
\includegraphics[width=\textwidth]{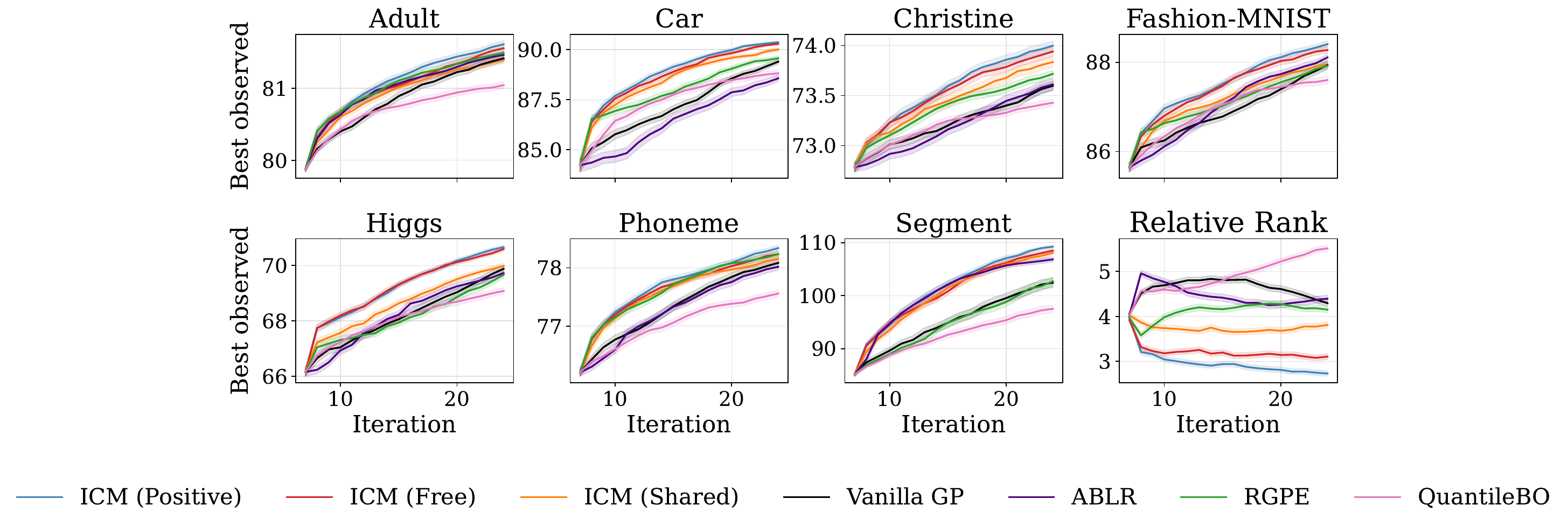}
\caption{\textbf{LCBench transfer, two positively correlated sources.} Best-observed target
validation accuracy vs.\ BO iteration for seven LCBench datasets (higher is
better); each target is warm-started with two positively correlated LCBench
source tasks, mean $\pm 1$~SE across $99$ seeds. The final panel shows each
method's relative rank aggregated across the seven datasets (lower is better).
ICM~(Positive) and ICM~(Free) reach higher best-observed accuracy and lower
average rank than the Vanilla GP and QuantileBO.}
\label{fig:lcbench_baseline}
\end{figure}

\paragraph{Reading.} Figure~\ref{fig:lcbench_baseline} plots, per dataset, the
best target validation accuracy observed so far versus BO iteration; the final
panel aggregates each method's per-seed rank across all seven datasets. The
per-task-standardized ICM~(Positive) and ICM~(Free) attain the lowest average
rank, ahead of the Vanilla GP and QuantileBO, with ICM~(Positive) best overall.
The textbook ICM~(Shared) lands mid-pack and does not match the
per-task-standardized variants, consistent with the standardization pitfall of
\S\ref{subsec:pitfall_standardize}. The appendix probes correlation inference
directly (App.\ Fig.\ \ref{fig:inferred_correlation}): when a source is
genuinely uninformative, the free-sign ICM~(Free) leaves the inferred correlation
diffuse and unidentified (the floor of Prop.~\ref{prop:cr_bound}), while the
sign-constrained ICM~(Positive) splits its mass between $0$ and $+1$ -- the
wrong-sign mode that Remedy~2 removes. Additional LCBench source configurations
and per-method ablations are in Fig.~\ref{fig:lcbench_sweepA},
\ref{fig:lcbench_sweepB}, \ref{fig:lcbench_baseline_abl},
\ref{fig:lcbench_sweepA_abl}, and \ref{fig:lcbench_sweepB_abl}.

\subsection{Surrogate-based HPO: pd1 and \textsc{ifeval}}
\label{subsec:hpo}

We complement the synthetic grid with surrogate-based HPO transfer whose task
structure is genuinely, rather than affinely, correlated. pd1~\citep{Wang2024}
contributes two deep-learning tuning targets, cifar10 and mnist, over a $4$-D
hyperparameter cube, each warm-started by a \emph{single highly correlated source}
(cifar10$\leftarrow$cifar100, mnist$\leftarrow$fashion) with $N_s{=}16$ source
observations (App.~\ref{app:task_pd1}). \textsc{ifeval}~\citep{Chen2024Admire}
scores LLM outputs against verifiable instruction-following constraints; the
optimization variable is a $19$-dimensional data-mixture simplex over
instruction-tuning sources used to fine-tune Qwen2.5, with two model-scale
source fidelities (0.5B and 3B) warm-starting the 7B target and $N_s{=}16$
source observations per fidelity. Since direct evaluation at every BO step is
infeasible, the optimized objective is a GP surrogate: a sample path of an MTGP
jointly fit to the real held-out data (App.~\ref{app:task_ifeval}). QuantileBO is
absent on \textsc{ifeval} (see the methods of \S\ref{sec:results}).

\begin{figure}[t]
\centering
\includegraphics[width=\textwidth]{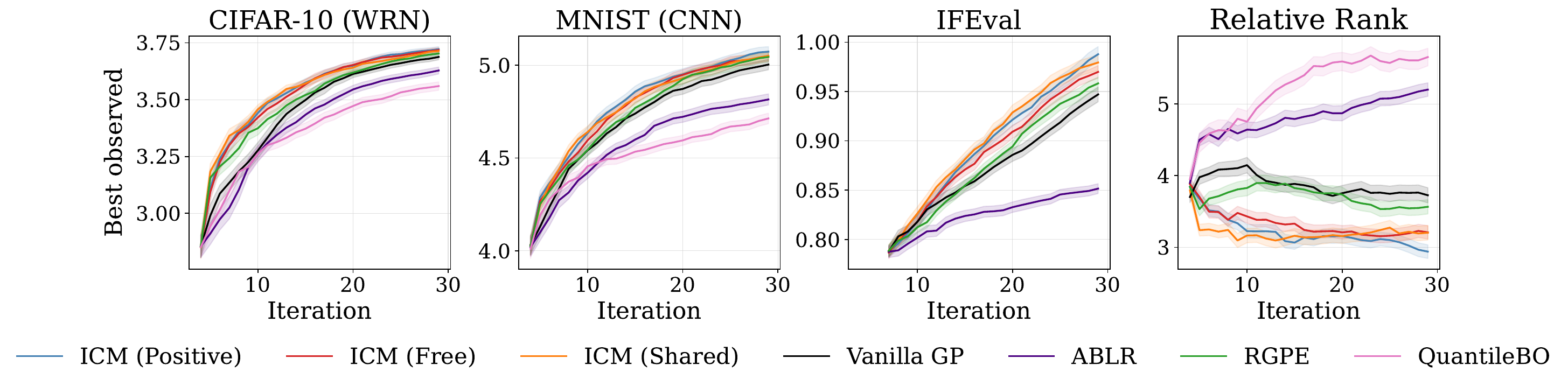}
\caption{\textbf{Surrogate-based HPO transfer: pd1 and \textsc{ifeval}.} Best-observed
target value vs.\ BO iteration for two pd1 targets (cifar10 (WRN), mnist (CNN))
and \textsc{ifeval}, with a final Relative Rank panel aggregating rank across the
three tasks (mean $\pm 1$~SE over $99$ seeds). pd1 uses one highly correlated
source per target; \textsc{ifeval} uses the 0.5B and 3B model-scale
sources. QuantileBO appears on the pd1 panels but not \textsc{ifeval} (see \S\ref{sec:results}). Per-target pd1 breakdowns are in
Fig.~\ref{fig:pd1}.}
\label{fig:hpo}
\end{figure}

\paragraph{Reading.} The combined view (Fig.~\ref{fig:hpo}) aggregates rank
across all three surrogate-based tasks and reads off whether the per-task-standardized
ICM variants retain their advantage on genuinely non-affine transfer.
ICM~(Positive) and ICM~(Free) each rank among the stronger configurations, with
no single configuration dominating throughout, consistent with the heterogeneity
expected on real HPO data; the per-target pd1 breakdowns appear in
Fig.~\ref{fig:pd1}.

\subsection{Effect of co-location}
\label{subsec:colocation}

We isolate Remedy~3 by varying how many of the $N_t = 8$ target observations on
Ackley-5 (a $5$-D instance, distinct from the grid's Ackley-6) and Hartmann-6 are
placed on randomly chosen source design points (Tab.~\ref{tab:colocation};
$g_t = f + h_t$, $\rho^\star = 0.80$, $N_s = 8$ source observations per task).
Without co-location the recovered correlation behaves close to a coin flip, with
frequent sign flips and $\pm 1$ corner saturation; co-locating more of the target
budget onto source design points steadily raises the mean recovered correlation
and sharply cuts the sign-flip rate, the empirical translation of
Prop.~\ref{prop:gp_dilution} with no model change. Notably, this improvement in
correlation inference holds even though the overlap points are chosen at random;
selecting the strongest source points for re-evaluation should improve it
further. The corresponding effect on optimization performance is inconclusive and
is reported in full in the appendix ablations (App.~\ref{app:extra_results}): the
source-design-overlap ICM variants (Fig.~\ref{fig:synthetic_grid_abl},
\ref{fig:lcbench_baseline_abl}, \ref{fig:lcbench_sweepA_abl},
\ref{fig:lcbench_sweepB_abl}) do not consistently beat their non-overlapping
counterparts, exposing the inference-versus-coverage trade-off of
\S\ref{subsec:pitfall_remedies}.

\begin{table}[t]
\centering
\small
\caption{\textbf{Effect of co-locating target observations on source-design
points.} Mean inferred task correlation $\hat\rho$, fraction of seeds where at
least one $\hat\rho_{0,t}$ has the wrong sign, fraction within $\pm 0.2$ of
$\rho^\star$ on both axes, and fraction saturating at a $\pm 1$ corner. True
$\rho^\star = 0.80$; $N_t = 8$, $N_s = 8$ per source; aggregated over $10$
seeds.}
\label{tab:colocation}
\begin{tabular}{llcccc}
\toprule
benchmark & $n_{\text{co}}$ & mean $\hat\rho$ & sign-flip frac. & within $\pm 0.2$ & $\pm 1$-saturation \\
\midrule
Ackley-5    & 0 & $+0.11$ & $0.70$ & $0.10$ & $0.20$ \\
            & 4 & $+0.47$ & $0.40$ & $0.40$ & $0.20$ \\
            & 8 & $+0.62$ & $0.10$ & $0.60$ & $0.00$ \\
\midrule
Hartmann-6  & 0 & $+0.15$ & $0.70$ & $0.30$ & $0.60$ \\
            & 4 & $+0.43$ & $0.40$ & $0.50$ & $0.30$ \\
            & 8 & $+0.49$ & $0.30$ & $0.50$ & $0.10$ \\
\bottomrule
\end{tabular}
\end{table}

\section{Limitations}
\label{sec:limitations}

This work isolates a previously underexplored failure mode of multi-task GP
transfer learning rather than fully solving it. Our real-world evaluation is
limited, and we do not claim a decisive solution: the three remedies are
conservative fixes for foundational problems in BOTL, restoring the Vanilla GP on
simple affine instances without closing the gap on harder instances or
alternative variants. The affine source-target family is a reasoning tool rather
than a restriction of the analysis: it is the simplest setting where the failure
modes are cleanly attributable. The same mechanisms apply more broadly; a fully
general analysis across non-affine task families is left to future work.

\section{Conclusions and Future Work}
\label{sec:conclusion}

We studied transfer learning in BO and isolated two structural pitfalls with
substantial practical impact on BOTL: finite-sample standardization noise that
propagates into the recovered task correlation
(Prop.~\ref{prop:standardize_propagation}), and an information-theoretic floor on
correlation inference (Prop.~\ref{prop:cr_bound}) that a GP at non-overlapping
designs further dilutes (Prop.~\ref{prop:gp_dilution}). We proposed three
conservative remedies -- per-task means and scales as model parameters, a
non-negativity constraint on the task correlation, and co-locating source and
target queries -- that address these problems and robustify BOTL more generally,
recovering the target-only baseline on the simple affine instances and improving
on existing transfer-learning methods in surrogate-based HPO transfer; on harder
instances and on most rank-based and latent-context variants the failure
persists. We also proposed an experimental setup that exposes these issues in a
clear, controlled way, which we hope future transfer-learning work can build on.

We believe that greater awareness of these pitfalls will enable more robust
transfer learning in BO. Better-conditioned priors on the task covariance,
identifiability-aware fitting procedures, and principled rules for declining
transfer when the source-side budget is too small to be informative remain open.

\clearpage
\bibliographystyle{plainnat}

\clearpage

\clearpage

\appendix

\section{Tasks}
\label{app:tasks}

\paragraph{Method set and naming.} The main figures use a lean method set:
Vanilla GP (target-only), ICM~(Shared), ICM~(Positive),
ICM~(Free), RGPE~\citep{Feurer2018},
ABLR~\citep{Perrone2018}, and QuantileBO~\citep{Salinas2020}
(the Gaussian-copula-prior method; dropped on \textsc{ifeval}, see \S\ref{sec:results}). The appendix ablation
figures (App.~\ref{app:extra_results}) show the ICM family only: the three base
ICM configurations (Shared, Positive, Free) plus the two source-design-overlap
ICM variants of Remedy~3, on the synthetic grid and LCBench. The three ICM
configurations differ only in their standardization and task-covariance settings:
\textbf{ICM~(Shared)} is the textbook configuration -- global standardization, a
single shared mean across tasks, and a non-negativity-constrained task covariance;
\textbf{ICM~(Positive)} uses per-task standardization (per-task means and scales,
Remedy~1) with a non-negativity constraint on the task correlations (Remedy~2);
\textbf{ICM~(Free)} keeps the per-task means and scales but frees the sign of the
task correlations. The two source-design-overlap variants (Remedy~3) are the
ICM~(Positive) and ICM~(Free) configurations with part of the target budget
re-evaluated on source design points. All methods run
$q\text{LogNEI}$ with the surrogate refit after every step; for MTGP variants the
acquisition pins the task column to the target.

\subsection{Synthetic affine-TL benchmark}
\label{app:task_synthetic}

\textbf{Domain.} Five standard BoTorch test functions used as the canonical target $f$:
Hartmann3~($d{=}3$), Hartmann6~($d{=}6$), Ackley~($d{=}6$), Levy~($d{=}4$), Levy~($d{=}5$).
Each function is evaluated at its natural dimension. Per-task signal scale $\sigma_f$
is the empirical standard deviation of $f$ on a Sobol probe.

\textbf{Source-task generation.} Per replication, $T{-}1$ source tasks are drawn as
positive-affine perturbations $g_t(\xv) = a_t\,\sigma_f\,f(\xv) + a_t\,\sigma_f\,b_t + \varepsilon_t$
with $a_t \sim \mathrm{LogNormal}(\mu{=}0.25,\sigma{=}0.5)$ (strictly positive),
$b_t \sim \mathcal{N}(0,1)$, and observation noise
$\varepsilon_t \sim \mathcal{N}(0, \sigma_{\mathrm{noise}}^2)$. The target is the
canonical $f$ ($a{=}1, b{=}0$). This is the affine family of \S\ref{sec:pitfall}
and Eq.~\eqref{eq:icm_joint_cov} rescaled by the per-task signal scale
$\sigma_f$, so slopes and offsets are expressed in $\sigma_f$ units. Source design
points are drawn from independent Sobol streams seeded by $\mathrm{seed}+1000+t$;
the configuration is summarized in Tab.~\ref{tab:synthetic_config}.

\begin{table}[h]
\centering
\small
\begin{tabular}{lll}
\toprule
Quantity & Symbol & Value \\
\midrule
Source slope & $a_t$ & $\mathrm{LogNormal}(0.25,\,0.5)$ \\
Source offset & $b_t$ & $\mathcal{N}(0,\,1)$ (in $\sigma_f$ units) \\
Source noise std & $\sigma_{\mathrm{noise}}$ & $0.1$ \\
Number of source tasks & $T-1$ & $2$ \\
Source observations per task & $N_s$ & $12$ \\
Target init points (Sobol) & $n_{\mathrm{init}}$ & $d+1$ \\
BO budget & $n_{\mathrm{BO}}$ & $30$ \\
Replications per (method, base fn) & seeds & $99$ \\
\bottomrule
\end{tabular}
\caption{Synthetic affine-TL benchmark configuration.}
\label{tab:synthetic_config}
\end{table}

\subsection{LCBench}
\label{app:task_lcbench}
\begin{figure}[t]
\centering
\includegraphics[width=\textwidth]{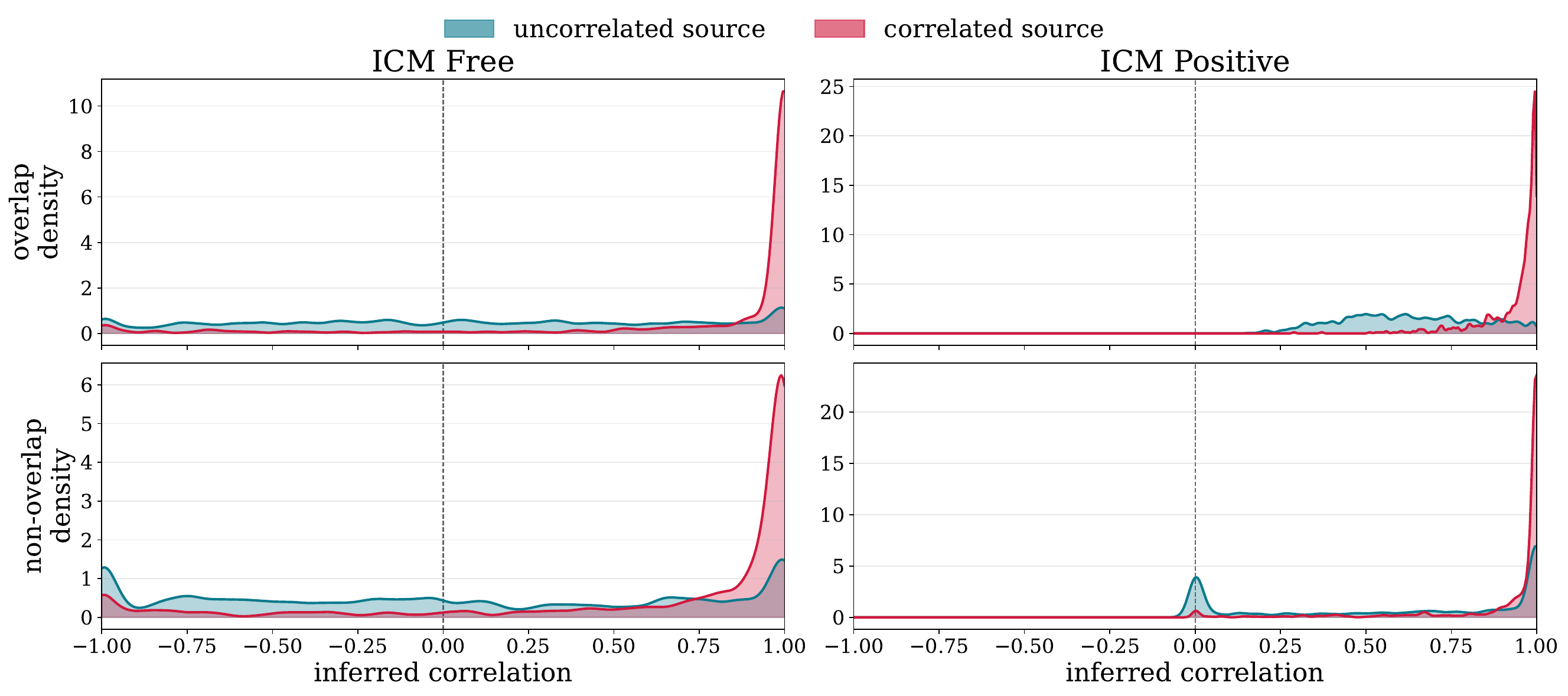}
\caption{\textbf{Inferred task correlation, correlated vs.\ uncorrelated source.}
Post-initialization density of the inferred target--source
correlation for ICM~(Free) and ICM~(Positive) (columns), with overlapping and
non-overlapping source designs (rows), for the one-correlated-plus-one-uncorrelated
source configuration; the crimson density is the
correlated source and the teal density the uncorrelated source. Against the
uncorrelated source, ICM~(Free) is \emph{diffuse and unidentified} while the
sign-constrained ICM~(Positive) is \emph{bimodal}, splitting mass between $0$ and
$+1$; both keep the genuinely correlated source concentrated near $+1$. This is
the direct evidence for the correlation-inference floor of
Prop.~\ref{prop:cr_bound} and the wrong-sign mode that Remedy~2 removes (see the
\S\ref{subsec:lcbench} \emph{Reading} text).}
\label{fig:inferred_correlation}
\end{figure}

LCBench~\citep{Zimmer2021} is used as a semi-synthetic benchmark: each target is
a GP surrogate fitted to a real classification dataset over a common $7$-D search
space, and its sources are small affine transformations of that surrogate. The
main figure (Fig.~\ref{fig:lcbench_baseline}) uses the \textbf{two-positive-source}
baseline: two sources, $12$ points per source, $\dim{=}7$. Targets are the \textbf{seven} datasets Car, Fashion-MNIST,
Higgs, Segment, Adult, Christine, and Phoneme (Numerai28.6 dropped). Budget is
$8$ init $+$ $17$ BO $=$ $25$ total, over $99$ seeds. Appendix-only variants use
the same budget/seeds: \textbf{one positive $+$ one uncorrelated source} (the
source configuration for the density panel of
Fig.~\ref{fig:inferred_correlation}), \textbf{one anti-correlated source},
and the per-method ablations of App.~\ref{app:extra_results}.

\subsection{pd1}
\label{app:task_pd1}
\textbf{Domain.} pd1~\citep{Wang2024} is a deep-learning hyperparameter-tuning
benchmark (non-affine) released as part of the HyperBO line of work; we optimize
a GP surrogate fit to its tabular data.
We use two targets, \{cifar10\_wrn, mnist\_simple\_cnn\}, over a
$4$-dimensional hyperparameter cube (learning rate, momentum, weight decay,
label-smoothing). Each target is warm-started by a \textbf{single
highly correlated source} (cifar10$\leftarrow$cifar100,
mnist$\leftarrow$fashion); see Tab.~\ref{tab:pd1_config}.

\begin{table}[h]
\centering
\small
\begin{tabular}{lll}
\toprule
Quantity & Symbol & Value \\
\midrule
Input dim & $d$ & $4$ \\
Targets & --- & cifar10\_wrn, mnist\_simple\_cnn \\
Source workloads per target & $T-1$ & $1$ (highly correlated) \\
Source obs.\ per workload & $N_s$ & $16$ \\
Target init (Sobol) & $n_{\mathrm{init}}$ & $d+1$ \\
BO budget & $n_{\mathrm{BO}}$ & $25$ \\
Replications & seeds & $99$ \\
\bottomrule
\end{tabular}
\caption{pd1 transfer benchmark configuration.}
\label{tab:pd1_config}
\end{table}

\subsection{\textsc{ifeval}, evaluated on a GP surrogate}
\label{app:task_ifeval}
\textbf{Domain.} 19-d data-mix simplex; the objective is the \textsc{ifeval} composite score for a
Qwen2.5 model finetuned on the chosen mixture. Sources: Qwen2.5 0.5B and 3B; target: 7B (Tab.~\ref{tab:ifeval_config}).

\textbf{Surrogate.} An MTGP is fit jointly on the real held-out data across all
three fidelities. A single multi-output posterior sample is drawn at $512$ Sobol
anchors (fixed seed) and interpolated per fidelity by a single-task GP, yielding
the GP surrogate that BO optimizes. The interpolation step works around missing
kernel feature generation on the product kernel; cross-fidelity correlation is
exact at the anchors.

\textbf{Source training data.} For each source fidelity, $N_s{=}16$ rows are uniformly randomly subsampled
from the raw observations (deterministic per seed). Source
$X,Y$ are real raw rows; only the BO target is a GP surrogate sample path. QuantileBO is
dropped on this benchmark (see \S\ref{sec:results}), leaving six methods.

\begin{table}[h]
\centering
\small
\begin{tabular}{lll}
\toprule
Quantity & Symbol & Value \\
\midrule
Input dim & $d$ & $19$ \\
Source fidelities & --- & Qwen2.5 0.5B, 3B \\
Target fidelity & --- & Qwen2.5 7B \\
Joint Sobol anchor count & --- & $512$ \\
Source obs.\ per fidelity & $N_s$ & $16$ \\
Target init (Sobol) & $n_{\mathrm{init}}$ & $8$ \\
BO budget & $n_{\mathrm{BO}}$ & $22$ \\
Replications & seeds & $99$ \\
Joint MTGP $\rho$ (analytical from $\Bm$) & --- & $\rho_{0.5,7}{=}0.48$, $\rho_{0.5,3}{=}0.76$, $\rho_{3,7}{=}0.94$ \\
\bottomrule
\end{tabular}
\caption{\textsc{ifeval} GP-surrogate benchmark configuration.}
\label{tab:ifeval_config}
\end{table}

\paragraph{Source construction for pd1 and \textsc{ifeval}.} For pd1 and
\textsc{ifeval} the optimized objective is a GP surrogate fit to \emph{real
held-out task data}: the objective surface is a sample path of a jointly fit MTGP
over the tabular data (a standard continuous-surrogate-over-tabular
construction), not a synthetic perturbation. LCBench, by contrast, fits a GP
surrogate to each real dataset and derives its sources as affine transformations
of that surrogate, making it semi-synthetic. The uncorrelated- and
anti-correlated-source configurations are used only on LCBench and the synthetic
grid; pd1 and \textsc{ifeval} use genuinely correlated tasks.

\paragraph{Reproducibility.}
Code and instructions for reproducing all experiments will be released as a public GitHub repository upon acceptance.

\section{Additional transfer-learning results}
\label{app:extra_results}

The main text references these LCBench variant and ablation figures
(\S\ref{subsec:lcbench}) and the standalone pd1 figures
(\S\ref{subsec:hpo}). All share the LCBench/HPO budgets and $99$ seeds of
App.~\ref{app:tasks}.

\begin{figure}[t]
\centering
\includegraphics[width=\textwidth]{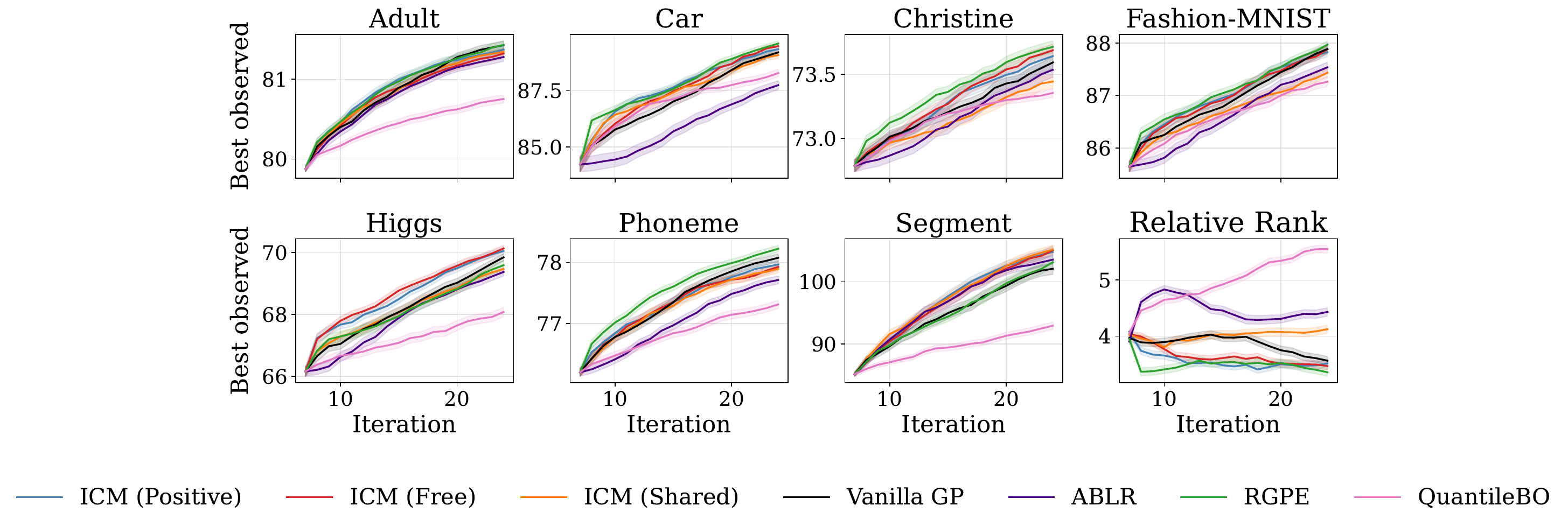}
\caption{\textbf{LCBench transfer, one positive $+$ one uncorrelated source.}
Best-observed target validation accuracy vs.\ BO iteration for the seven LCBench
datasets (higher is better), with a final relative-rank panel (lower is better);
mean $\pm 1$~SE across $99$ seeds. Each target is warm-started by one genuinely
correlated and one uncorrelated source. The per-task-standardized ICM~(Positive)
and ICM~(Free) retain their rank advantage despite the uninformative second
source; this is the source configuration from which the correlation-density
diagnostic of Fig.~\ref{fig:inferred_correlation} is drawn.}
\label{fig:lcbench_sweepA}
\end{figure}

\begin{figure}[t]
\centering
\includegraphics[width=\textwidth]{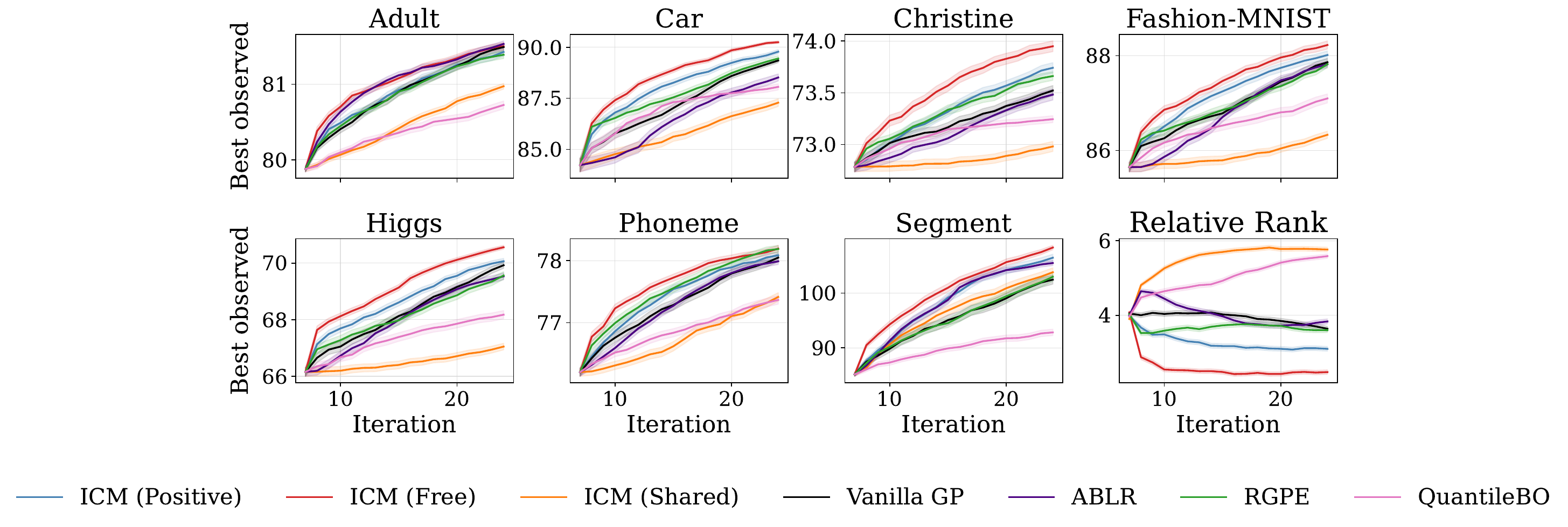}
\caption{\textbf{LCBench transfer, one anti-correlated source.} Best-observed
target validation accuracy vs.\ BO iteration for the seven LCBench datasets
(higher is better), with a final relative-rank panel (lower is better); mean
$\pm 1$~SE across $99$ seeds. Each target is warm-started by a single
anti-correlated source -- the negative-transfer stress test for Remedy~3
(\S\ref{subsec:pitfall_remedies}). The sign-constrained ICM~(Positive) is the
most robust to the misleading source, while free-sign methods are more exposed
to negative transfer.}
\label{fig:lcbench_sweepB}
\end{figure}

\begin{figure}[t]
\centering
\includegraphics[width=\textwidth]{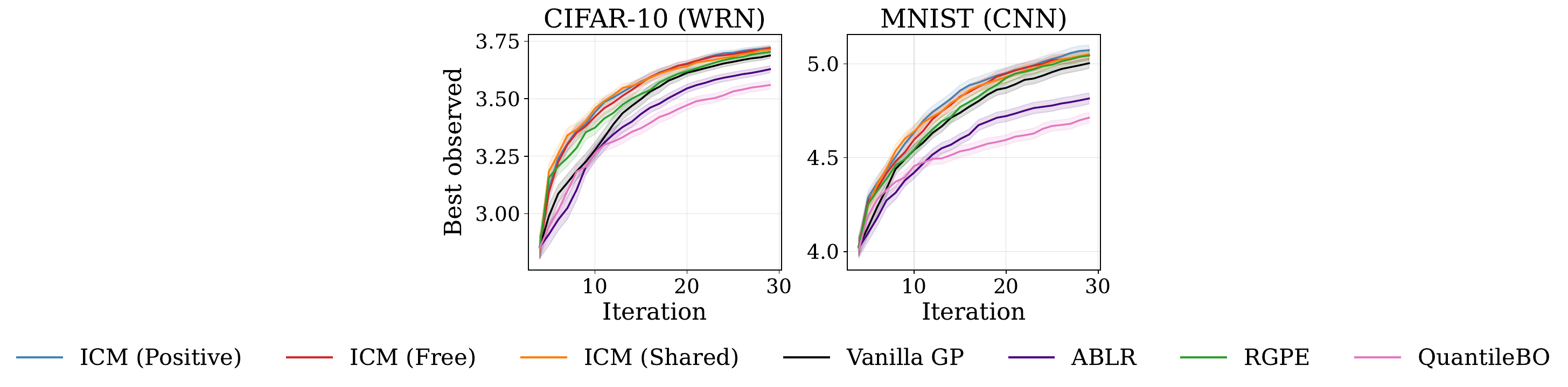}
\caption{\textbf{Standalone pd1 per-target breakdowns.} Best-observed target
value vs.\ BO iteration for the cifar10 and mnist targets separately, each
warm-started by one highly correlated source; mean $\pm 1$~SE over $99$ seeds.
The per-target view resolves the aggregate of Fig.~\ref{fig:hpo}: the
per-task-standardized ICM~(Positive) and ICM~(Free) are competitive on both
targets, with no single method dominating across the two.}
\label{fig:pd1}
\end{figure}

\begin{figure}[t]
\centering
\includegraphics[width=\textwidth]{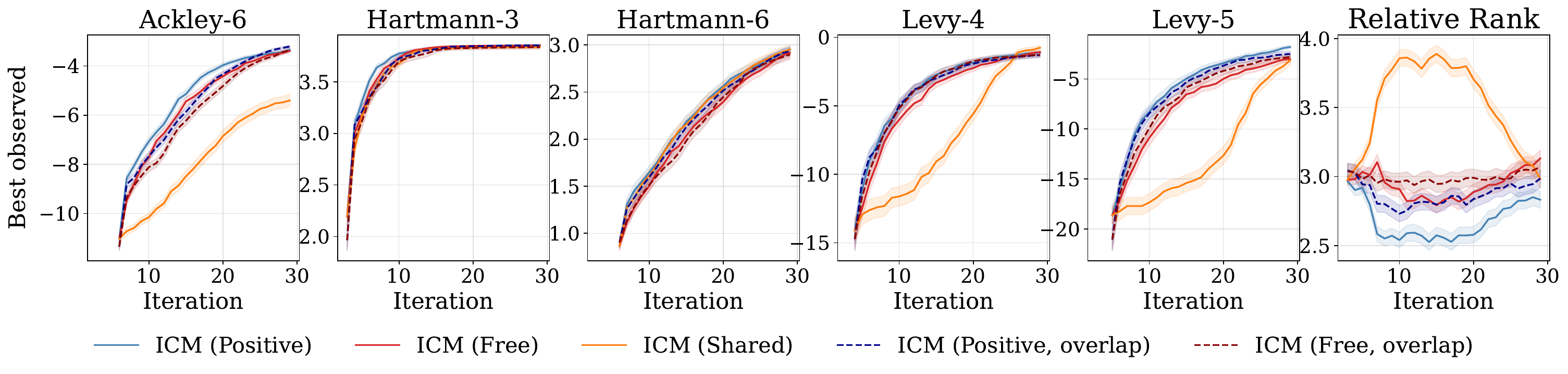}
\caption{\textbf{Synthetic grid, per-method ablation.} Best-observed target value
vs.\ acquisition iteration on the five synthetic base functions, mean $\pm 1$~SE
across seeds, with a final relative-rank panel (lower is better). This ablates
the headline grid (Fig.~\ref{fig:synthetic_grid}) over the ICM family only,
adding the two source-design-overlap variants (Remedy~3) omitted from the main
figure. The overlap variants do not consistently beat their non-overlapping
counterparts, exposing the inference-versus-coverage trade-off of
\S\ref{subsec:pitfall_remedies}.}
\label{fig:synthetic_grid_abl}
\end{figure}

\begin{figure}[t]
\centering
\includegraphics[width=\textwidth]{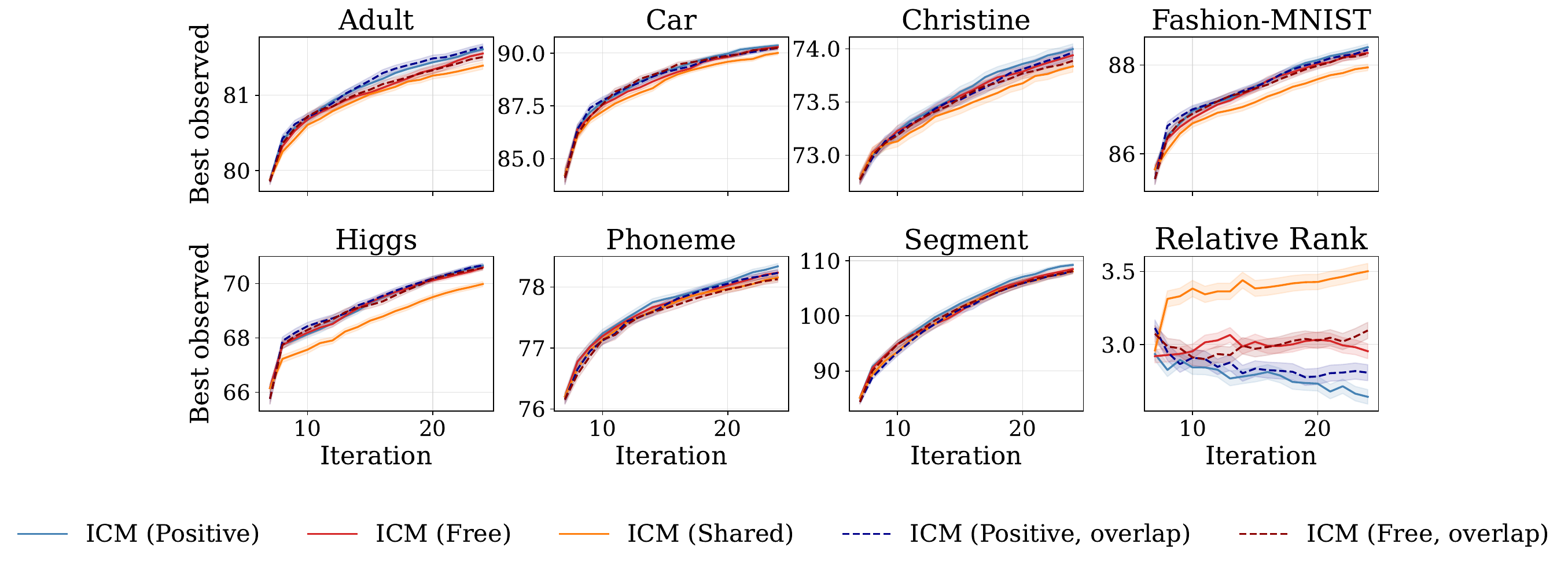}
\caption{\textbf{LCBench two-positive-source baseline, per-method ablation.}
Best-observed target validation accuracy vs.\ BO iteration across the seven
LCBench datasets, with a final relative-rank panel; mean $\pm 1$~SE over $99$
seeds. This ablates the baseline of Fig.~\ref{fig:lcbench_baseline} over the ICM
family, adding the source-design-overlap variants (Remedy~3), which track their
non-overlapping counterparts closely.}
\label{fig:lcbench_baseline_abl}
\end{figure}

\begin{figure}[t]
\centering
\includegraphics[width=\textwidth]{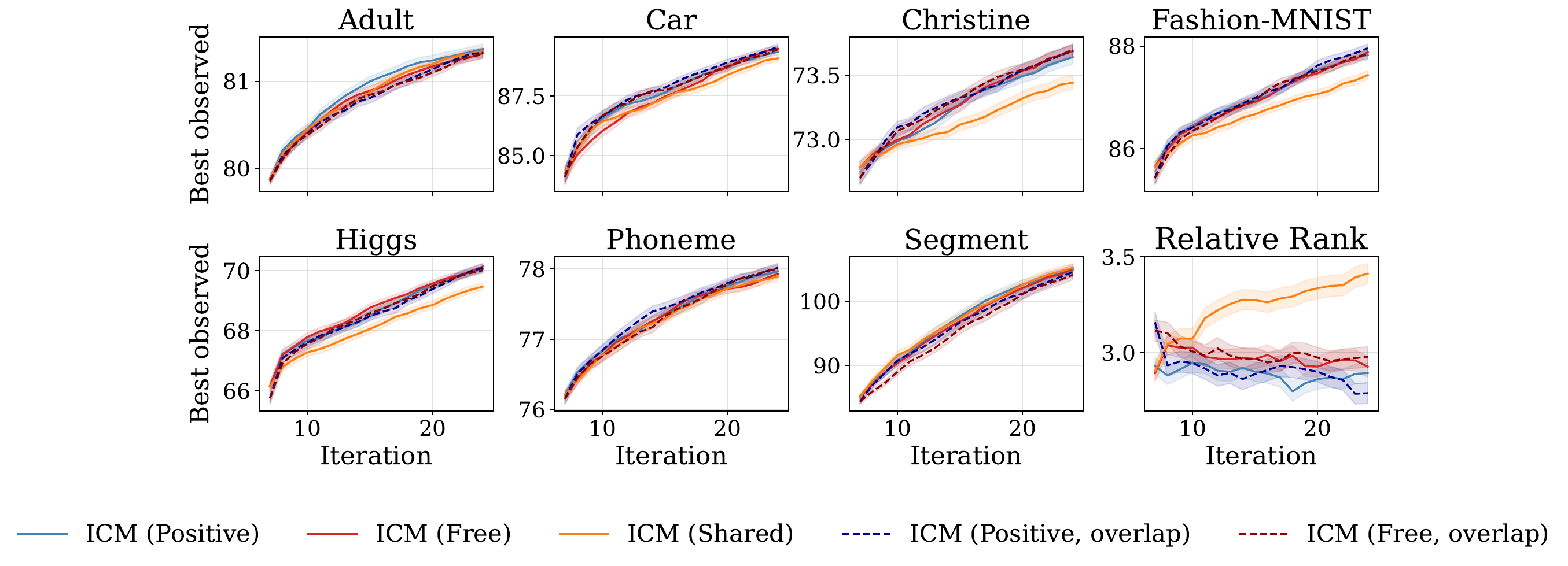}
\caption{\textbf{LCBench one-positive-plus-one-uncorrelated-source, per-method
ablation.} Best-observed target validation accuracy vs.\ BO iteration across the
seven LCBench datasets, with a final relative-rank panel; mean $\pm 1$~SE over
$99$ seeds. This ablates Fig.~\ref{fig:lcbench_sweepA} over the ICM family,
adding the source-design-overlap variants (Remedy~3); co-location does not
consistently improve optimization performance despite sharpening correlation
inference.}
\label{fig:lcbench_sweepA_abl}
\end{figure}

\begin{figure}[t]
\centering
\includegraphics[width=\textwidth]{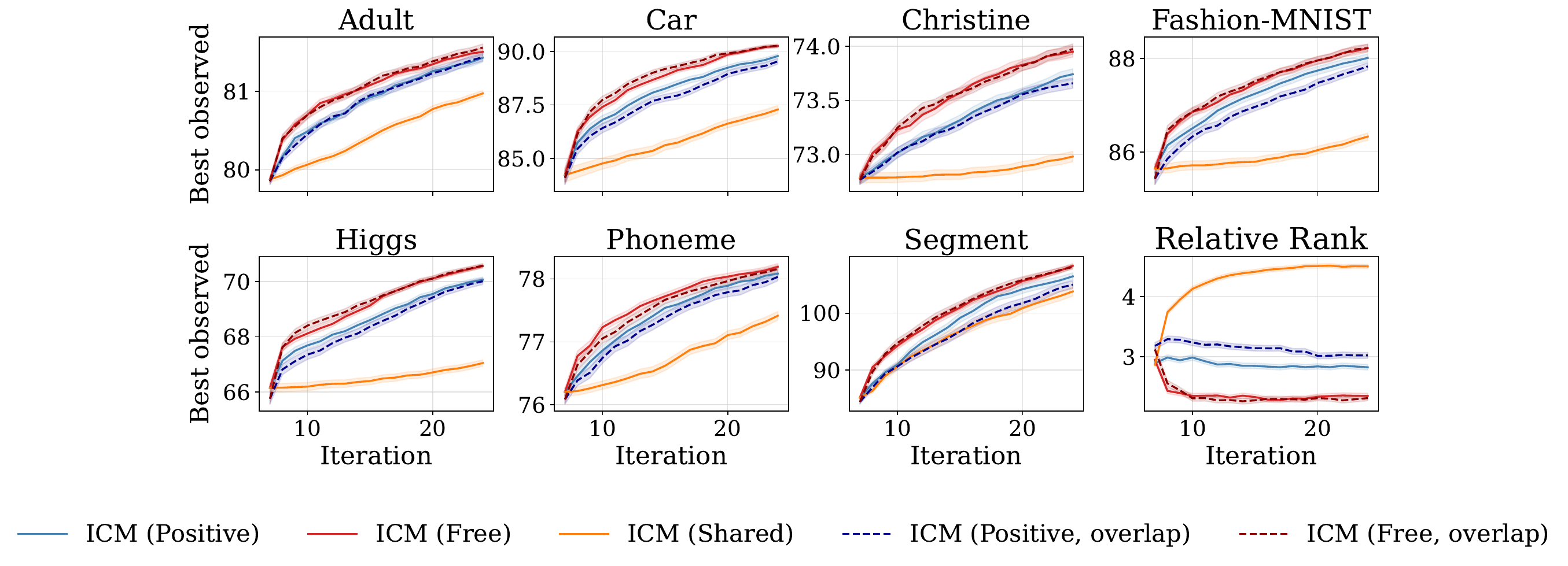}
\caption{\textbf{LCBench one-anti-correlated-source, per-method ablation.}
Best-observed target validation accuracy vs.\ BO iteration across the seven
LCBench datasets, with a final relative-rank panel; mean $\pm 1$~SE over $99$
seeds. This ablates Fig.~\ref{fig:lcbench_sweepB} over the ICM family, adding the
source-design-overlap variants (Remedy~3); the sign-constrained ICM~(Positive)
remains the most robust to the anti-correlated source.}
\label{fig:lcbench_sweepB_abl}
\end{figure}

\section{Full proof of Proposition~\ref{prop:standardize_propagation}}
\label{app:standardize_harm_proof}

Let $y_1,\dots,y_{N_s}$ be the source observations with empirical mean $\hat\mu_s$
and empirical variance $\hat\sigma_s^2 = N_s^{-1}\sum_i(y_i-\hat\mu_s)^2$. We work
under $\rho^\star_{st}=1$ and treat the target divisor $\hat\sigma_t$ as fixed at
its truth, since the target is data-rich with $N_t \gg N_s$.

\paragraph{Step 1: sampling error of the standardization estimates.}
For Gaussian, or more generally finite-fourth-moment, observations the central
limit theorem gives
\begin{align}
    \hat\mu_s &= \mu_s + \mathcal{N}\!\bigl(0,\tfrac{\sigma_s^2}{N_s}\bigr) + o_p(N_s^{-1/2}),\\
    \hat\sigma_s^2 &= \sigma_s^2 + \mathcal{N}\!\bigl(0,\tfrac{2\sigma_s^4}{N_s}\bigr) + o_p(N_s^{-1/2}).
\end{align}
The delta method with $g(u)=\sqrt{u}$ and $g'(\sigma_s^2)=1/(2\sigma_s)$ then gives
\begin{equation}
    \hat\sigma_s = \sigma_s + \mathcal{N}\!\bigl(0,\tfrac{\sigma_s^2}{2N_s}\bigr) + o_p(N_s^{-1/2}).
\end{equation}

\paragraph{Step 2: propagation into the task covariance.}
Per-task standardization divides the source by $\hat\sigma_s$ instead of
$\sigma_s$, scaling its row and column of the task covariance by
$\sigma_s/\hat\sigma_s$. Under $\rho^\star_{st}=1$ the true blocks are
$B^\star_{st}=\sigma_s\sigma_t$, $B^\star_{ss}=\sigma_s^2$, and
$B^\star_{tt}=\sigma_t^2$, so the standardized blocks are
\begin{equation}
    B^{\mathrm{eff}}_{st} = \frac{\sigma_s}{\hat\sigma_s}\,B^\star_{st},
    \qquad
    B^{\mathrm{eff}}_{ss} = \frac{\sigma_s^2}{\hat\sigma_s^2}\,B^\star_{ss},
    \qquad
    B^{\mathrm{eff}}_{tt} = B^\star_{tt}.
\end{equation}

\paragraph{Step 3: residual variance in $\hat\rho_{st}$.}
Write $\eta_s := (\hat\sigma_s-\sigma_s)/\sigma_s$, so that by Step~1
$\eta_s = \mathcal{N}(0,\tfrac{1}{2N_s}) + o_p(N_s^{-1/2})$. The recovered
correlation is
\begin{equation}
    \hat\rho_{st}
    = \frac{B^{\mathrm{eff}}_{st}}{\sqrt{B^{\mathrm{eff}}_{ss}\,B^{\mathrm{eff}}_{tt}}}
    = \frac{(1+\eta_s)^{-1}B^\star_{st}}{\sqrt{(1+\eta_s)^{-2}\sigma_s^2\;\sigma_t^2}}
    = \rho^\star_{st},
\end{equation}
so the scale factor cancels to first order and the bias is second order. The
surviving first-order stochastic term is the standardization error $\eta_s$
itself, propagated jointly with the mean estimate $\hat\mu_s$ that enters
$\hat\sigma_s^2$ through the same $N_s$ observations, giving
\begin{equation}
    \frac{\hat\rho_{st}}{\rho^\star_{st}}
    = 1 + \mathcal{N}\!\bigl(0,\tfrac{1}{2N_s}\bigr) + o_p(N_s^{-1/2}).
\end{equation}
The standard error is $\Theta(1/\sqrt{N_s})$ regardless of $N_t$: the source-side
sample size dominates. At $N_s=1$ the convention $\hat\sigma_s=1$ makes the
divisor pure noise; because the target posterior weights the source by
$\hat\rho_{st}$, this noise enters target predictions directly and can drive the
multi-task posterior strictly below the single-task baseline. \qed

\section{Full proof of Proposition~\ref{prop:cr_bound}}
\label{app:cr_bound_proof}

\paragraph{Step 1: eigenmode decomposition.}
Consider the two-task ICM model under the affine setup with shared design
$X = \{x_1, \ldots, x_N\}$ and observations $y_t = f_t(X) + \varepsilon_t$,
$\varepsilon_t \sim \mathcal{N}(0, \sigma^2 \Iv_N)$, and suppose all
hyperparameters except $\rho := B_{12}/\sqrt{B_{11}B_{22}}$ are known. Stacking
$\yv = (\yv_1, \yv_2) \in \R^{2N}$ gives
$\bm{\Sigma} = \Bm \otimes \Km_x + \sigma^2 \Iv_{2N}$. Diagonalizing
$\Km_x = \bm{U} \bm{\Lambda} \bm{U}^\top$ and applying $\Iv_2 \otimes \bm{U}^\top$
splits the joint law into $N$ independent bivariate Gaussians, mode $i$ having
covariance $\widetilde{\bm{\Sigma}}_i = \lambda_i \Bm + \sigma^2 \Iv_2$.

\paragraph{Step 2: per-mode Fisher information.}
Define the effective SNR of mode $i$,
\begin{equation}
    u_i := \frac{\lambda_i^2 B_{11} B_{22}}{(\lambda_i B_{11} + \sigma^2)(\lambda_i B_{22} + \sigma^2)} \in [0, 1].
\end{equation}
Direct computation of
$\tfrac12 \mathrm{tr}\bigl((\widetilde{\bm{\Sigma}}_i^{-1}
\partial_\rho \widetilde{\bm{\Sigma}}_i)^2\bigr)$ gives
\begin{equation}
    I_i(\rho) = \frac{u_i(1 + u_i \rho^2)}{(1 - u_i \rho^2)^2},
\end{equation}
which increases in $u_i$, with $u_i = 1$ exactly in the noiseless limit
$\sigma^2 = 0$.

\paragraph{Step 3: aggregate bound.}
Summing the per-mode informations,
\begin{equation}
    I_N(\rho) = \sum_i I_i(\rho) \le N\,\frac{1+\rho^2}{(1-\rho^2)^2},
\end{equation}
and the Cram\'er--Rao bound yields
$\Var[\hat\rho] \ge (1-\rho^2)^2/[N(1+\rho^2)]$, with equality in the noiseless
limit. The rate stays $\Theta(1/N)$ at any fixed noise level; only the constant
degrades, controlled by the smallest eigenmode of $\Km_x$. For mismatched designs
with $|X_s| = N_s \le |X_t| = N_t$, a kernel-overlap argument gives
$I(\rho) \le c\, N_s (1+\rho^2)/(1-\rho^2)^2$, so the smaller task dominates the
rate.

\paragraph{Step 4: Spearman asymptotic relative efficiency.}
At the bivariate normal the Pearson MLE attains the Cram\'er--Rao bound
$\Var[\hat\rho] = (1-\rho^2)^2/N$ in the noiseless limit. The Spearman rank
estimator $\hat\rho_S$ has the asymptotic variance derived by
\citet{Hoeffding1948} and tabulated by \citet{Borkowf2002}, whose ratio to the
Pearson variance is
\begin{equation}
    \frac{\Var[\hat\rho_S]}{\Var[\hat\rho]} = \frac{\pi^2}{9} \approx 1.097,
    \qquad\text{i.e.\ an asymptotic relative efficiency (ARE) of }
    \frac{9}{\pi^2} \approx 0.912.
\end{equation}
The Spearman standard error therefore inherits the same $\Theta(1/\sqrt N)$ rate,
inflated by at most $10\%$. The rank weights of RGPE and the copula kernel of
QuantileBO are smooth functionals of the empirical rank vector and inherit this
rate through the functional delta method.

\section{Full proof of Proposition~\ref{prop:gp_dilution}}
\label{app:gp_dilution_proof}

\paragraph{Step 1: packing the design.}
Let $k_x$ be a stationary correlation kernel on $\Omega \subset \R^d$ with
characteristic length $\ell$, so $k_x(x,x') \le \exp(-\|x-x'\|/\ell)$ up to a
kernel-dependent constant. Let $X = \{x_1,\dots,x_N\} \subset \Omega$ be the
design and $X_\ell \subseteq X$ a maximal subset with pairwise distance
$\ge \ell$. Volume comparison gives $|X_\ell| \le (\mathrm{diam}(\Omega)/\ell)^d$.

\paragraph{Step 2: effective-rank bound.}
Points inside a common packing cell satisfy $|k_x(x_i,x_j)| \ge \exp(-1)$, so
$\Km_x$ is within $O(1)$ of a block-constant matrix with $|X_\ell|$ blocks. Up to
a fixed constant, $\Km_x$ therefore has effective rank
$N_{\mathrm{eff}} := |X_\ell| \le \min\{N,\,(\mathrm{diam}(\Omega)/\ell)^d\}$, and
only the top $N_{\mathrm{eff}}$ eigenmodes carry signal.

\paragraph{Step 3: Fisher information.}
Reusing the eigenmode decomposition of App.~\ref{app:cr_bound_proof}, the Fisher
information for $\rho$ is $I_N(\rho) = \sum_i I_i(\rho)$ with $I_i$ proportional to
the SNR $u_i$ of mode $i$. Modes outside the packing have $u_i = O(\exp(-2))$ and
contribute $O(1)$ in total, so the sum is dominated by the $N_{\mathrm{eff}}$
in-packing modes:
\begin{equation}
    I_N(\rho) \;\le\; N_{\mathrm{eff}}\,\frac{1+\rho^2}{(1-\rho^2)^2}
    \;\le\; \min\!\bigl\{N,\,(\mathrm{diam}(\Omega)/\ell)^d\bigr\}\,\frac{1+\rho^2}{(1-\rho^2)^2}.
\end{equation}
The Cram\'er--Rao bound then gives
$\Var[\hat\rho] \ge (1-\rho^2)^2/[N_{\mathrm{eff}}(1+\rho^2)]$: the GP identifies
$\rho$ from at most $N_{\mathrm{eff}}$ effective paired observations. \qed

\end{document}